\documentclass[lettersize,journal]{IEEEtran}
\usepackage{amsmath,amsfonts}
\usepackage{algorithmic}
\usepackage{algorithm}
\usepackage{array}
\usepackage[caption=false,font=normalsize,labelfont=sf,textfont=sf]{subfig}
\usepackage{textcomp}
\usepackage{stfloats}
\usepackage{url}
\usepackage{verbatim}
\usepackage{graphicx}
\usepackage{cite}
\usepackage{xcolor}
\usepackage{arydshln}
\usepackage{booktabs}
\usepackage{amsmath}
\usepackage{subcaption}

\begin{document}




\title{Exploration of Marker-Based Approaches in Argument Mining through Augmented Natural Language\thanks{This paper has been accepted for publication at the IEEE/INNS International Joint Conference on Neural Networks (IJCNN) 2025. © 2025 IEEE. Personal use of this material is permitted. Permission from IEEE must be obtained for all other uses.}}





\author{
\IEEEauthorblockN{Nilmadhab Das, Vishal Choudhary, V. Vijaya Saradhi, Ashish Anand}

\IEEEauthorblockA{Applied Machine Learning (AMaL) Lab} 

\IEEEauthorblockA{Department of Computer Science and Engineering, Indian Institute of Technology (IIT), Guwahati, India}

\IEEEauthorblockA{\{nilmadhabdas, v.choudhary, saradhi, anand.ashish\}@iitg.ac.in}
}

\maketitle 

\begin{abstract}
Argument Mining (AM) involves identifying and extracting  \textit{Argumentative Components (ACs)} and their corresponding \textit{Argumentative Relations (ARs)}. Most of the prior works have broken down these tasks into multiple sub-tasks. Existing end-to-end setups primarily use the dependency parsing approach. This work introduces a generative paradigm-based end-to-end framework \textit{arg}TANL.  \textit{arg}TANL frames the argumentative structures into label-augmented text, called \textit{Augmented Natural Language (ANL)}. This framework jointly extracts both ACs and ARs from a given argumentative text. Additionally, this study explores the impact of \textit{Argumentative} and \textit{Discourse} markers on enhancing the model's performance within the proposed framework. Two distinct frameworks, Marker-Enhanced \textit{arg}TANL (\textit{ME-arg}TANL) and \textit{arg}TANL with specialized Marker-Based Fine-Tuning, are proposed to achieve this. Extensive experiments are conducted on three standard AM benchmarks to demonstrate the superior performance of the \textit{ME-arg}TANL.
\end{abstract}

\begin{IEEEkeywords}
Argument Mining, Marker, Augmented Natural Language.
\end{IEEEkeywords}

\section{Introduction}
\label{sec:intro}
\IEEEPARstart{A}{rgument Mining (AM)} \cite{lawrence_argument_2019} deals with the detection and classification of ACs and their corresponding ARs from discourse dynamics. 
AM task has been subdivided into four sub-tasks \cite{niculae_argument_2017}: \textit{(i) Component Segmentation} entails identifying Argumentative Discourse Units (ADUs) \cite{peldszus_ranking_2013}, \textit{(ii) Component Classification} involves categorizing ADUs into various ACs \cite{feng_classifying_2011}, \textit{(iii) Relation Identification} focuses on detecting argumentative relationships among two or more ADUs \cite{carstens_towards_2015}, and \textit{(iv) Relation Classification} deals with classifying these identified relations into different ARs \cite{jo-etal-2021-classifying}. However, following the study in \cite {morio_end--end_2022}, we collectively term the initial pair of sub-tasks as Argument Component Extraction (ACE) and the subsequent pair as Argumentative Relation Classification (ARC). Consequently, end-to-end AM, referred to as ACRE in this work, involves jointly addressing both ACE and ARC tasks. AM is useful for various downstream tasks such as debate analysis \cite{Lawrence2017UsingAS}, automated essay scoring \cite{Nguyen2018ArgumentMF}, customer review analysis \cite{chen-etal-2022-argument}, etc.

The primary challenge in AM lies in effectively handling the longer sequence length of ACs and their associated ARs \cite{lawrence_argument_2019}. Defining boundaries for ACs is more intricate compared to tasks like Named Entity Recognition (NER) or Parts-of-Speech (POS) tagging, where the target text span consists of a few tokens only. Also, every AC has certain underlying contexts of argumentativeness and is related to another AC of the same context. Variations in argument representations across domains pose another challenge \cite{Daxenberger2017WhatIT}. Previous approaches to this task have primarily utilized extractive methods \cite{he_generative_nodate} to address these challenges, which have shown limited performance. Recently, end-to-end frameworks based on dependency parsing \cite{morio_end--end_2022} have gained popularity due to their ability to streamline the extraction process into a single unified model. However, these methods often require intricate post-processing steps, adding to the overall complexity of the solution. Given the limitations of these recent approaches, we propose an alternative method based on a text-to-text generation paradigm. This new approach reduces complexity by eliminating the need for elaborate post-processing. 

Several generative frameworks have recently been proposed for various NLP tasks, including NER \cite{yan-etal-2021-unified-generative}, joint entity and relation extraction \cite{zaratiana2024autoregressive}, and co-reference resolution \cite{quan-etal-2019-gecor}. One such framework is Translation between Augmented Natural Language (TANL)\cite{tanl}, which addresses a range of NLP tasks within a unified framework. This framework generates outputs by augmenting the original input text with class labels specific to downstream tasks. ``\textit{[ Tolkien $|$ \textbf{person} ]’s epic novel [ The Lord of the Rings $|$ \textbf{book} ] was published in 1954-1955.}" Here, labels such as \textit{person} and \textit{book} are embedded within the original text. With this kind of output format, complex post-processing steps are reduced. However, among all the NLP tasks solved by the TANL framework, the target output spans (e.g., \textit{Tolkien} or \textit{The Lord of the Rings}) are relatively shorter as compared to the target spans of AM tasks, where each AC spans across longer sections of text. Thus, it is a meaningful avenue to explore the capability of TANL in solving AM tasks that consist of longer target spans.

Studies in \cite{gao_discourse_2022} have demonstrated that markers are strongly correlated with improvements in AM task performance as they reliably indicate the presence of ACs in the text. Two distinct types of markers are categorized: \textit{(a) Argumentative Markers} and \textit{(b) Discourse Markers (DMs)}.
Argumentative Markers are typically phrases such as \textit{``I strongly agree that,” ``But, I deny the point that,” or ``However, this clearly proves that,”} conveying the argumentativeness of the discourse. In contrast, DMs are single-token connectives like \textit{``But,” ``And,” or ``However,”} representing the rhetorical structure of language. The set of argumentative markers is extracted using the rule-based methods by \cite{kuribayashi_empirical_2019}, while the set of DMs is constructed from a specific web-corpus by \cite{sileo-etal-2020-discsense}. However, rule-based extraction yields a large number of irrelevant spans lacking generalizability. We employ a similar extraction method for \textit{Argumentative Markers}, incorporating an additional manual filtering step to eliminate irrelevant or non-generic spans.

\begin{figure}
  \centering
  \includegraphics[width=0.45\textwidth]{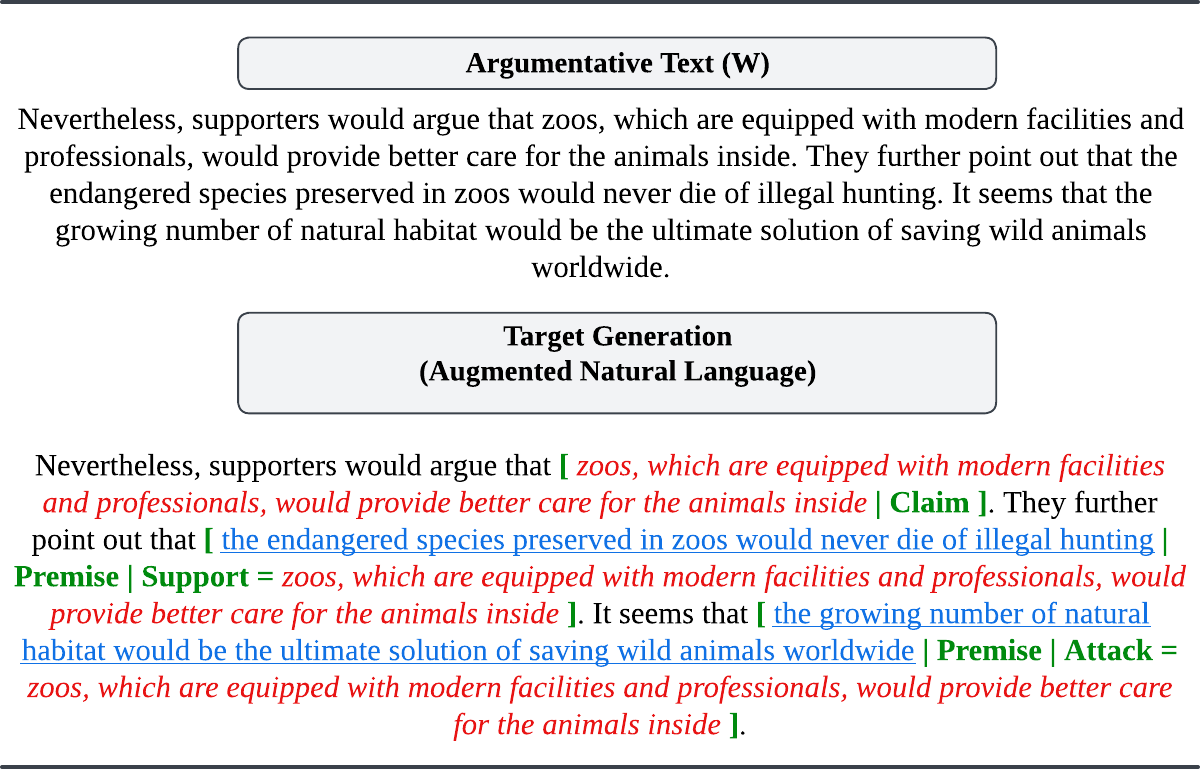}
  \caption{An overview of the proposed generative end-to-end argument mining framework \textbf{\textit{arg}TANL}. Claims are highlighted in \textit{italics} and marked in red. Premises are \underline {underlined} and marked in blue. Augmented labels are represented with \textbf{bold} font and marked in green.}
  \label{fig:am_example}
\end{figure}

We propose the following. (1) \textbf{\textit{arg}TANL}, a TANL framework for the downstream task of computational AM. In this, both ACs and ARs are label-augmented to form an ANL-based generation sequence (See Figure \ref{fig:am_example}). This framework handles longer target AC spans than the target spans of previously attempted downstream tasks. (2) \textbf{Marker-Based Fine-Tuning of \textit{arg}TANL} to enhance its performance. From the different marker sets \textit{(e.g., Argumentative Markers and DMs)}, the marker information is learned first with specifically designed fine-tuning strategies. Then with this learned marker information, the \textbf{\textit{arg}TANL} output is learned. (3) \textbf{Marker Enhanced \textit{arg}TANL (\textit{ME-arg}TANL)}, an enhanced \textit{arg}TANL framework by incorporating the marker knowledge inside the output text itself. This is distinctly different from the TANL framework, where in addition to the class labels of ACs and ARs, the marker information is also included in the output. \textit{ME-arg}TANL allows to perform two tasks jointly: (a) Marker Identification and (b) AC and AR identification. With this framework, the model receives a strong signal from the markers, which are present just before the AC, to guide the identification of that AC inside the same generation sequence. This, in turn, enhances the overall performance of the AM task. Whenever the marker information is present inside the AC span the \textit{ME-arg}TANL framework is redundant, In that case, the Marker-Based Fine-Tuning framework is appropriate.

We conduct a comprehensive set of experiments to evaluate the performance of the proposed framework, organized into several subsections. We first analyze the main results with a detailed comparison of \textit{arg}TANL, Marker-Based Fine-Tuned \textit{arg}TANL, and \textit{ME-arg}TANL to highlight the improvements brought by the knowledge of markers. Then we compare the \textit{Component-Only and End-to-End} variants to determine their relative effectiveness. Next, we analyze the \textit{Impact of input length} on the performance of the \textit{arg}TANL. Then, we conduct an ablation study by introducing \textbf{Abbreviated \textit{arg}TANL}. Here, repeated full AC spans are replaced with unique ``ID" tokens, reducing output sequence length and improving efficiency in handling longer paragraphs. This study explores the performance trade-offs between the standard and abbreviated formulations. Finally, an \textit{Error Analysis} is performed to find the scope for further improvement. Through these extensive experiments upon three standard benchmarks of AM literature, the proposed approach achieves competitive results in several datasets and surpasses various important baselines. In summary, the main contributions of this paper are:


\begin{enumerate}
    \item A generative task formulation in the form of \textbf{\textit{arg}TANL} framework to solve \textit{End-to-End AM} along with \textit{Component-only} variant.
    \item We enhance \textit{arg}TANL with a specialized \textit{\textbf{Marker-Based Fine-Tuning}} by leveraging the knowledge of markers to improve the AM task performance.
    \item We propose \textbf{\textit{ME-arg}TANL}, an advanced version of \textit{arg}TANL that integrates marker information directly into the output sequence. It identifies markers and argumentative structures jointly to deliver superior AM task performance.
    \item An extensive performance analysis along with a target-oriented ablation study with \textbf{Abbreviated \textit{arg}TANL} to justify the strength of the proposed formulation.
\end{enumerate}

\section{Related Work}
\label{sec:relwrk}
\noindent\textbf{Argument Mining:} Prior studies largely focus on subsets of the four AM sub-tasks. Recent works \cite{ye-teufel-2021-end, he_generative_nodate} emphasize joint formulations in an end-to-end manner. \cite{persing_end--end_2016} adopted a pipelined approach for ACE and ARC, optimizing error propagation via Integer Linear Programming (ILP). \cite{eger_neural_2017} reformulated the end-to-end AM task into sequence tagging, dependency parsing, multi-task tagging, and relation extraction. \cite{ye-teufel-2021-end} proposed a biaffine dependency parsing approach for AM. Dataset scarcity was highlighted by \cite{morio_end--end_2022}, who introduced a cross-corpora multi-task span-biaffine framework, employing a span classifier with BIO tagging and span representations via average pooling. \cite{he_generative_nodate} framed AM as a \textit{text-to-sequence} generation task, producing sequences of AC and AR types with start/end indices. Recently, the TANL framework was used by \cite{kawarada-etal-2024-argument} to address end-to-end AM as text-to-text generation, but exploration of markers and ANL formats remained limited. Our work addresses this gap.

\noindent\textbf{Markers:} Markers are pivotal signals for ADUs \cite{dutta-etal-2022-unsupervised}. \cite{stab_parsing_2017} used markers as lexical features for classifying argument components via multiclass classification. \cite{kuribayashi_empirical_2019} extracted 1131 \textit{argumentative markers} to improve span representations in AM tasks. \cite{dutta-etal-2022-unsupervised} explored markers in Reddit threads, extracting 69 Reddit-specific markers and applying \textit{selective masked language modeling (sMLM)} for domain adaptation. The model was then used for \textit{Argument Component Identification}, with marker-like tokens predicted to infer \textit{Relation Types}. However, marker exploration for AM in a generative, end-to-end framework remains unexplored.

\noindent\textbf{Augmented Natural Language (ANL) \& Generative Paradigm:} Generative methods are reframing NLP tasks as generation problems, with label-augmented text (\textit{ANL}) being a notable strategy. ANL has been applied to NER \cite{athiwaratkun_augmented_2020}, sentiment analysis \cite{zhang_towards_2021}, and relation extraction \cite{liu_autoregressive_2022}. \cite{tanl} employed ANL for structured prediction tasks such as joint entity and relation extraction, event argument extraction, and coreference resolution by framing them as text-to-text generation problems.

\section{Task Formulation}
\label{sec:task}

We represent argumentative text as $W=w_1,w_2,w_3,....,w_n$, where $n$ is the total number of tokens in $W$. For a text-span $w_i, w_{i+1}, w_{i+2}, \ldots, w_{j}$ in $W$, we write it as $w_{i:j}$. We define a set of AC types as $T^c=\{t_1^c, t_2^c, t_3^c,....,t_{n_c}^c \}$ and a set of AR types as $T^r=\{t_1^r, t_2^r, t_3^r,....,t_{n_r}^r \}$, where $n_c$ and $n_r$ refer to a total number of possible AC and AR types respectively. We use a set of symbol tokens, $S = \{[, ],=,|\}$ in the ANL, where $``["$ marks the start of a component, $``]"$ marks the end of a component, $``="$ is used for relation assignments, and $``|"$ is used to separate labels. Subsequent sections discuss different formulations of the proposed task.

\subsection{ACE task: Component-only Variant}
\label{sec:ACE}
For any given argumentative text $w_{1:n}$, objective of the \emph{ACE} is
to extract a set of ACs as $C=\{C_i|C_i=(c_i,c_i^s,c_i^e)\}$, where $C_i$ is the $i^{th}$ AC, $c_i \in T^c$, and $c_i^s$ and $c_i^e$ refer to the relative start and end token indices of $c_i$ respective to $W$. Here, we generate the label-augmented text for ACs only such as, in Figure \ref{fig:am_example}, for the head, AC is \textit{[ the endangered species preserved in zoos would never die of illegal hunting $|$ Premise ]} and for tail AC is \textit{[ zoos, which are equipped with modern facilities and professionals, would provide better care for the animals inside $|$ Claim ]}. The main motivation for this variant is to assess our proposed method's ability to identify ACs alone when the relational information is muted.

\begin{figure}
  \centering
  \includegraphics[width=0.4\textwidth]{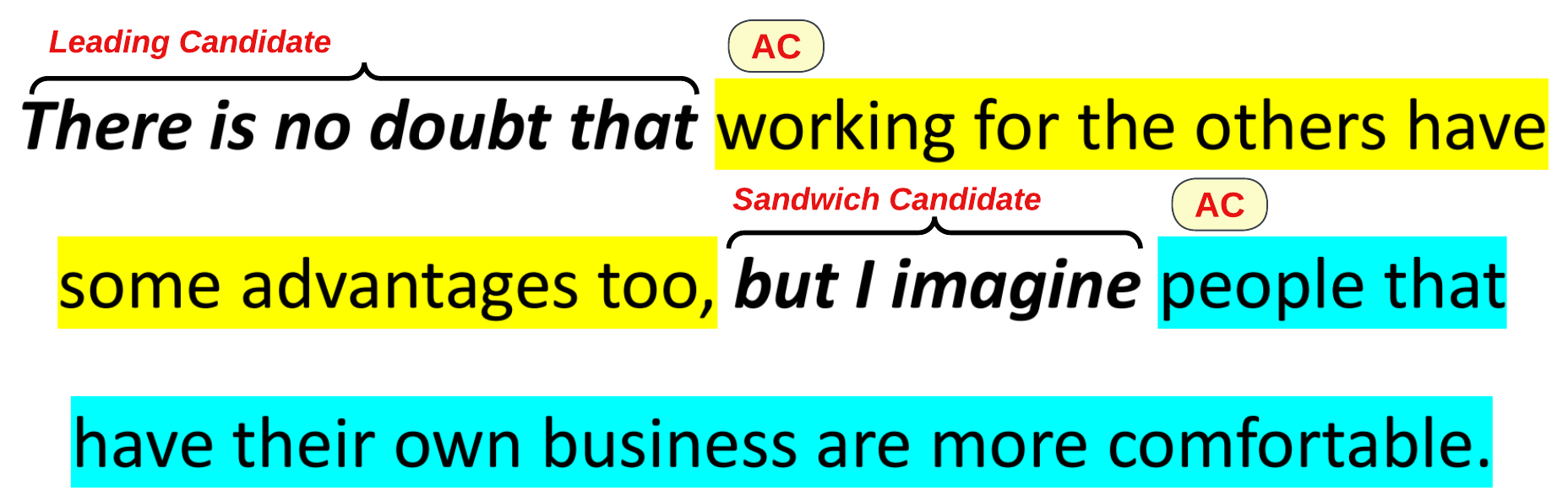}
  \caption{An example sentence from \texttt{AAE} corpus describing both ways of marker extraction. Here, extracted \textbf{\textit{marker candidates}} are in bold italics, and ACs are highlighted.}
  \label{fig:marker_extraction_steps}
\end{figure}



\subsection{ACRE task: End-To-End Argument Mining}
\label{sec:acre}
The proposed end-to-end formulation jointly frames the \emph{ACE} and \emph{ARC} tasks in the following manner. We define ARs as $R=\{R_i|R_i=(r_i, r_i^h, r_i^t)\}$, where $R_i$ is the $i^{th}$ AR corresponding to AR type $r_i \in T^r$, and $r_i^h$ and $r_i^t$ refer to the head and tail ACs respectively. $r_i^h$ and $r_i^t$ are connected with relation type $r_i$. If two components $C_p, C_q \in C$ are related with $R_k \in R$ as $r_k^h = C_p$ and $r_k^t = C_q$, then for the token spans of $C_p$ i.e. $w_{c_p^s:c_p^e}$ and $C_q$ i.e. $w_{c_q^s:c_q^e}$, the model will generate augmented labels as $[w_{c_p^s:c_p^e} | c_p | r_k = w_{c_q^s:c_q^e}]$ and $[w_{c_q^s:c_q^e} | c_q ]$ respectively. The rest of the tokens in $W$ will be rewritten as it is. We refer to this joint formulation as \emph{ACRE}. Figure \ref{fig:am_example} illustrates the \emph{ACRE} formulation.

\section{Methodology}
\label{sec:method} 


In this section, we provide an overview of the \textit{arg}TANL framework, the marker extraction process, the marker-based fine-tuning approach, and the \textit{ME-arg}TANL framework.

\subsection{The \textbf{\textit{arg}TANL} Framework}
\label{sec:argtanl}

This framework adopts the TANL formulation (Figure \ref{fig:am_example}) with specific customization tailored for the AM task. Given argumentative text as input, the framework generates ANL consisting of ACs and ARs as label-augmented texts enclosed with symbol tokens, as described in Section \ref{sec:task}. This approach leverages the strengths of the TANL framework while introducing modifications to effectively capture the complex relationships and structures inherent in argumentative texts. The integration of symbolic tokens ensures precise boundary definition and clear distinction between different ACs and ARs, facilitating accurate identification within the generated sequences.



\begin{table*}
\caption{Description of the initial step of \textbf{\textit{Marker-Based Fine-Tuning}} of the \textit{arg}TANL framework.}
\centering
\renewcommand{\arraystretch}{1.15}
\begin{tabular}{p{1.3cm} p{6.6cm} p{7.5cm}}
\hline
\textbf{Strategy} & \textbf{Input Sequence} & \textbf{Target Generation Sequence}\\
\hline
    A-MKT & Last but not least, students have ... difficulties.  & \textbf{[ Last but not least, $|$ marker ]} students have ... difficulties.\\

\hline
    SM-MKT & \textit{$<$extra\_id\_0$>$ $<$extra\_id\_1$>$ $<$extra\_id\_2$>$ $<$extra\_id\_3$>$ $<$extra\_id\_4$>$} students have ... difficulties.
    & \textit{$<$extra\_id\_0$>$} \textbf{Last} \textit{$<$extra\_id\_1$>$} \textbf{but} \textit{$<$extra\_id\_2$>$} \textbf{not} \textit{$<$extra\_id\_3$>$} \textbf{least} \textit{$<$extra\_id\_4$>$} \textbf{,} \textit{$<$extra\_id\_5$>$} \\
\hline
    E-MKT &  Last but not least, students have ... difficulties. & [\textbf{-1,-1,-1,-1,-1,}0,0,0,0,0,0,0]\\
\hline
    D-MKT & Motivations for playing cricket are vastly different. It is a well-crafted game.
    & Motivations for playing cricket are vastly different. \textbf{Truly,} it is a well-crafted game.\\
\hline
\end{tabular}
\label{strategies}

\end{table*}

\subsection{Argumentative Marker Extraction}
\label{sec:marker_ex}
An argumentative marker typically signals the beginning of an AC. However, not every marker is always followed by an AC, and an AC may not always be preceded by a marker. Considering this phenomenon, we extract two types of (See Figure \ref{fig:marker_extraction_steps}) potential marker candidates from any argumentative text: \textit{(i) Leading candidates:} by extracting tokens from the start of a sentence to the beginning of an AC, and \textit{(ii) Sandwich candidates:} by extracting tokens after the end of an AC until the start of another AC in the same sentence. However, these strategies yielded some spans, which were non-generic. For example, ``\textit{In spite of the importance of \textbf{sports activities}}" or ``\textit{Nevertheless, opponents of \textbf{online-degrees} would argue that}" are having non-generic spans. These spans lacked generalizability across diverse topics, being tailored solely to their respective subjects. To enhance generalizability, we introduce an additional layer of manual filtering to eliminate topic-dependent spans from the pool of extracted marker candidates. Initially, we identified 2925 marker candidates. After manual filtering and removing duplicates, we refine the list to 1072 unique \textit{argumentative markers}.

\subsection{Marker-Based Fine-Tuning of \textit{arg}TANL}
\label{sec:marker-based-tine-tuning}

This technique strategically divides the \textit{arg}TANL output generation into two significant steps. The initial step leverages the extracted markers from various corpora to execute Marker-Based Fine-Tuning. This involves implementing four distinct generative fine-tuning strategies, each utilizing varied input and output combinations (see Table \ref{strategies}). The objective is to acquaint the model with the nuanced representations of markers within the argumentative text. Notably, neither the markers from the test data nor the test data itself were used during this initial fine-tuning process.

In the second step, the models derived from the initial step undergo additional fine-tuning to perform the \textit{arg}TANL task. This way, we get four unique resultant models, each distinguished by the applied fine-tuning strategy. Below are the details of different Marker-Based Fine-Tuning strategies performed as the first step of this technique:

\subsubsection{Augmented Marker Knowledge Transfer (A-MKT)}
This strategy takes plain text input and fine-tunes the model to generate ANL where only \textit{markers} are augmented. ACs and ARs are not augmented. 

\subsubsection{Span-Masked Marker Knowledge Transfer (SM-MKT)}
It is a self-supervised denoising fine-tuning strategy by \textit{masking the span of markers}. We replace every token in the marker span with sentinel tokens. Here, the target generation sequence is formed by concatenating the sentinel tokens and the corresponding marker tokens.

\subsubsection{Marker Knowledge Transfer through Encoding (E-MKT)}
Unlike the above strategies, which are based upon text-to-text generation, it is a \textit{text-to-sequence} generation strategy. Here, we generate the labels of marker tokens in terms of a numeric sequence of 0's and -1's, where -1 and 0 are replacing the markers and non-markers of the input text, respectively.

\subsubsection{Discourse Marker Knowledge Transfer (D-MKT)}
To check the effectiveness of single-token DMs over multiple-token markers, we propose this fine-tuning strategy. Using sentence pairs from a standard DM corpus, we generate a target sequence in the following format: \textit{(Sentence 1 + DM + Sentence 2)}, where the concatenated input is \textit{(Sentence 1 + Sentence 2)}. Thus the model can achieve the capability of generating the probable \textit{``connective"} between a pair of sentences.

\begin{figure}
  \centering
  \includegraphics[width=0.35\textwidth]{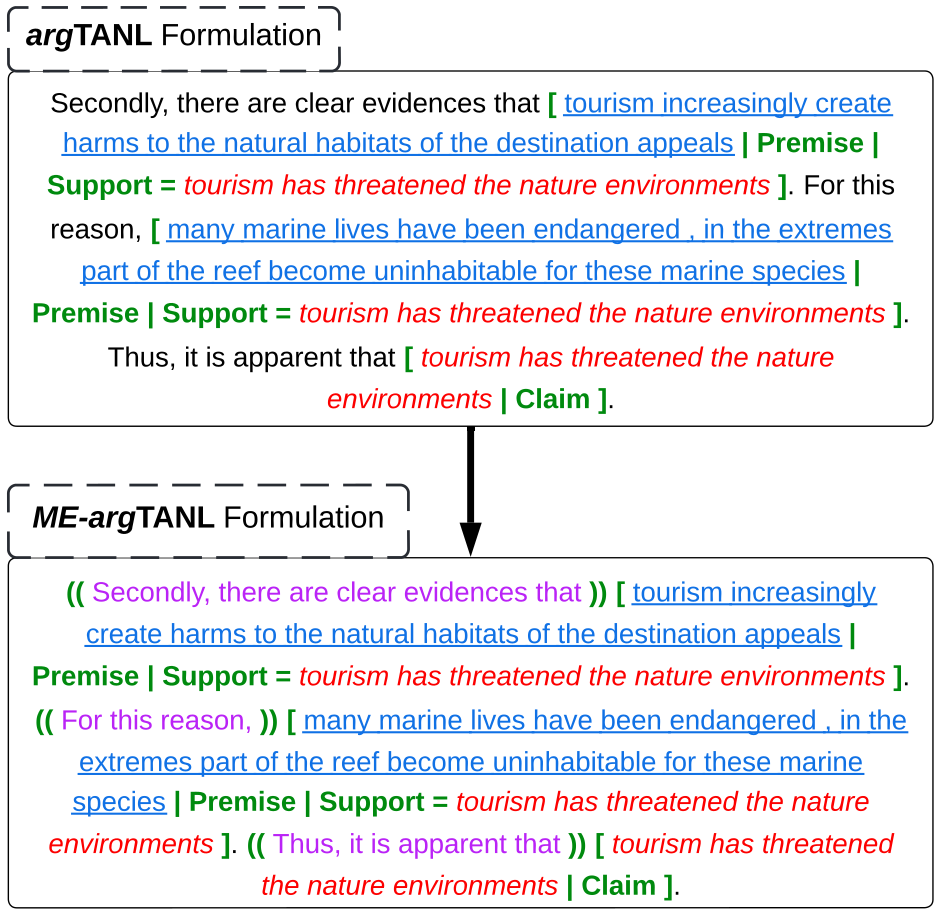}
  \caption{Description of \textbf{\textit{ME-arg}TANL} Formulation. Markers are enclosed with the ``(( ))" symbol in violet color. Claims are highlighted in red \textit{italics}. Premises are \underline {underlined} in blue. Augmented labels are represented in green with \textbf{bold} font.}
  \label{fig:marker_augmented}
\end{figure}

\subsection{The \textit{ME-arg}TANL Framework}
\label{sec:meargtanl}

This framework modifies the \textit{arg}TANL to improve the performance of AM tasks by uniquely embedding marker knowledge into the same generation sequence. We include the marker span using \textit{``(( ))"} to form an unique marker span as \textit{``(( Marker ))"}  (See Figure \ref{fig:marker_augmented}). These double braces serve as a special enclosing symbol, indicating the task difference between the \textit{arg}TANL task, and the task of marker identification. Given that marker identification occurs within the same target sequence and the generation process is autoregressive, with markers appearing mostly on the left side of the target AC span, there is a strong likelihood that the augmented markers will be generated first. Subsequently, these markers will serve as a guiding signal for the generation of ACs and ARs, which are present immediately afterward. This type of sequential generation will enrich the model by engaging it in two types of structurally independent tasks simultaneously.

\section{Experimental Setup}
\subsection{Datasets}
We consider three standard datasets to evaluate the proposed methods as follows:




\noindent \textbf{Argument Annotated Essay (AAE)}\cite{stab_parsing_2017}: This dataset contains 402 student essays annotated at the segment (span) level. Every essay is divided into multiple paragraphs. A total of 1833 paragraphs are annotated with three AC types, $T^c = \{\text{\textit{Claim, MajorClaim, Premise}}\}$, and two AR types, $T^r =\{\text{\textit{Support, Attack}}\}$. \texttt{AAE} contains a large number of argumentative markers, with almost every AC beginning with one. We extract argumentative markers only from this dataset and transfer this knowledge to all experiments.

\noindent \textbf{Fine-Grained Argument Annotated Essay (AAE-FG)}\cite{schaefer-etal-2023-towards}: Recently, the \texttt{AAE} dataset is annotated in a more fine-grained manner by \cite{schaefer-etal-2023-towards}. The $\text{\textit{Major Claim}}$ and $\text{\textit{Claim}}$ classes are divided into $\text{\textit{Fact}}$, $\text{\textit{Value}}$, $\text{\textit{Policy}}$. And the $\text{\textit{Premise}}$ class is divided into $\text{\textit{Common}}$ $\text{\textit{Ground}}$, $\text{\textit{Testimony}}$, $\text{\textit{Hypothetical}}$ $\text{\textit{Instance}}$, $\text{\textit{Statistics}}$, $\text{\textit{Real}}$ $\text{\textit{Example}}$ and $\text{\textit{Others}}$. So, it contains a total of nine AC types and the ARs remain unchanged.

\noindent \textbf{Consumer Debt Collection Practices (CDCP)}\cite{niculae_argument_2017}: This dataset contains 731 user comments collected from the Consumer Financial Protection Bureau (CFPB) website. It contains five AC types, $T^c = \{\text{\textit{Fact, Testimony, Reference, Policy, Value}}\}$, and two AR types, $T^r = \{\text{\textit{Reason, Evidence}}\}$.

We use the \texttt{Discovery} \cite{sileo-etal-2020-discsense} corpus as the primary source of DMs. It is used for the sole purpose of the initial fine-tuning of the \textit{Marker-Based Fine-Tuning} only for its D-MKT strategy. Extracted from the \textit{DepCC} web corpus \cite{panchenko-etal-2018-building}, it features 1.74 million pairs of adjacent sentences \textit{(Sen1, Sen2)} with 174 DMs, consolidating 10k pairs per DM. All DMs occur at the beginning of \textit{Sen2}.

Notably, the \textit{arg}TANL framework and the \textit{Marker-Based Fine-Tuning} of \textit{arg}TANL are designed to be applicable across all datasets. Whereas the \textit{ME-arg}TANL framework is only designed for both versions of the \texttt{AAE} corpus as this is the only dataset that contains markers outside the component boundary. The \texttt{CDCP} dataset includes the DMs inside the AC span boundaries. So, the formation of the \textit{ME-arg}TANL output is not trivially possible for this dataset.

\begin{table}
\caption{Experiment results on ACRE task with comparable baselines. Best scores are marked in \textbf{bold}. * indicates the baseline results we produced by running their open-source code.}
\label{baselines}
\centering
\renewcommand{\arraystretch}{1.15}
\scriptsize
\begin{tabular}{lllcc}
\hline
\textbf{Corpus} & \textbf{Model} & \textbf{Params} & \multicolumn{2}{c}{\textbf{Micro-F1}} \\
 & & & \textbf{ACE} & \textbf{ARC} \\
 
\hline

 & BiPAM & 110M & 41.15 & 10.34 \\
 & BART-B & 139M & 56.15 & 13.76 \\
 & RPE-CPM & 139M & 57.72 & 16.57 \\
 & CPM-only & 139M & 58.13 & 15.11 \\
 & Span-Biaffine-ST & 149M & \textbf{68.90} & \textbf{31.94}\\
 CDCP & SP-AM \textit{(Flan-T5-Base)} & 220M & 66.80 & 23.19\\
 & $T5_{argTANL}$ \textbf{(Ours)} & 220M & 66.56 & 19.81 \\
 & $T5_{E-MKT}$ \textbf{(Ours)} & 220M & 57.31 & 7.97 \\
 & $T5_{A-MKT}$ \textbf{(Ours)} & 220M & 61.06 & 13.86 \\
 & $T5_{SM-MKT}$ \textbf{(Ours)} & 220M & 64.17 & 16.68 \\
 & $T5_{D-MKT}$ \textbf{(Ours)} & 220M & 67.49 & 20.68 \\

\hline
 & Span-Biaffine-ST* & 149M & 59.54 & \textbf{54.75}\\
 & $T5_{argTANL}$ \textbf{(Ours)} & 220M & 60.41 & 32.14 \\
 & $T5_{E-MKT}$ \textbf{(Ours)} & 220M & 55.24 & 25.83 \\
 AAE-FG & $T5_{A-MKT}$ \textbf{(Ours)} & 220M & 57.66 & 27.82 \\
 & $T5_{SM-MKT}$ \textbf{(Ours)} & 220M & 59.37 & 30.54 \\
 & $T5_{D-MKT}$ \textbf{(Ours)} & 220M & 59.71 & 31.03 \\
 & $T5_{ME-argTANL}$ \textbf{(Ours)} & 220M & \textbf{61.39} & 33.12 \\

\hline

 & LSTM-Parser & - & 58.86 & 35.63 \\
 & ILP & - & 62.61 & 34.74 \\
 & BLCC & - & 66.69 & 39.83 \\
 & LSTM-ER & - & 70.83 & 45.52 \\
 & BiPAM & 110M & 72.90 & 45.90 \\
 & BiPAM-Syn & 110M & 73.50 & 46.40 \\
 & BART-B & 139M & 73.61 & 47.93 \\
 AAE & RPE-CPM & 139M & 75.94 & 50.08 \\
 & SP-AM \textit{(Flan-T5-Base)} & 220M & 75.55 & 58.51\\
 & DENIM & 139M & 76.50 & 58.51\\
 & Span-Biaffine-ST & 149M & 76.48 & \textbf{59.55}\\
 & $T5_{argTANL}$ \textbf{(Ours)} & 220M & 75.93 & 50.56 \\
 & $T5_{E-MKT}$ \textbf{(Ours)} & 220M & 73.06 & 45.89 \\
 & $T5_{A-MKT}$ \textbf{(Ours)} & 220M & 74.22 & 48.01 \\
 & $T5_{SM-MKT}$ \textbf{(Ours)} & 220M & 75.91 & 49.08 \\
 & $T5_{D-MKT}$ \textbf{(Ours)} & 220M & 76.45 & 49.91 \\
 & $T5_{ME-argTANL}$ \textbf{(Ours)} & 220M & \textbf{77.14} & 52.12 \\

 \hline




\end{tabular}
\end{table}

\subsection{Training Details}

We conducted all experiments using the \textit{T5-Base} model \cite{raffel2020exploring} on Nvidia A100 GPUs, with each reported result averaged over 5 runs\footnote{The implementation will be made available upon acceptance.}. The hyperparameter settings for the \textit{arg}TANL framework are as follows: For the \texttt{CDCP} dataset, a batch size of 4 is used, while for \texttt{AAE} and \texttt{AAE-FG}, the batch size is set to 8. Across all datasets, we utilize the AdamW optimizer with a learning rate of 0.0005. The maximum input length is set to 1024 tokens for \texttt{CDCP} and 512 tokens for \texttt{AAE} and \texttt{AAE-FG}. In the end-to-end and component-only training scenarios, we run for more than 10000 steps, with checkpoints saved every 200 steps.
During inference, beam search is used with a beam length of 8 across all datasets. For all strategies (A-MKT, SM-MKT, D-MKT, and E-MKT) of Marker-Based Fine-Tuning, we use a batch size of 16, the AdamW optimizer with a learning rate of 0.0005, a maximum input length of 512 tokens, and 10 epochs of training.
We use the following libraries: (i) TANL framework\footnote{\url{https://github.com/amazon-science/tanl}}, and (ii) HuggingFace's Transformers\footnote{\url{https://github.com/huggingface}}.

\subsection{Baseline Methods}
We consider several important baselines to investigate the efficacy of our end-to-end AM formulation on each dataset as follows: 

\begin{enumerate}
    \item \textbf{ILP} \cite{persing_end--end_2016}: Rich feature based approach to perform \textit{joint inference} over the AM sub-tasks optimized by\textit{ Integer Linear Programming (ILP)}.
    \item \textbf{BLCC} \cite{eger_neural_2017}: Based upon \textit{Bi-LSTM-CNN-CRF (BLCC)} to formulate this task as a sequence tagging problem.
    \item \textbf{LSTM-ER} \cite{eger_neural_2017}: An adapted version of an end-to-end relation extraction model with sequential LSTM \cite{miwa-bansal-2016-end}.
    \item \textbf{LSTM-Parser} \cite{eger_neural_2017}: A dependency parsing approach built on stacked LSTM \cite{dyer-etal-2015-transition}.
    \item \textbf{BiPAM} \cite{ye-teufel-2021-end}: Another dependency parsing approach with customized biaffine operation based upon BERT-base \cite{devlin-etal-2019-bert}.
    \item \textbf{BiPAM-syn} \cite{ye-teufel-2021-end}: An enhanced version of BiPAM with the inclusion of \textit{syntactic} information.
    
    \item \textbf{BART-B} \cite{he_generative_nodate}: A generative approach to \textit{text-to-sequence generation} with Bidirectional and Auto-Regressive Transformer (BART) \cite{lewis-etal-2020-bart}.
    \item \textbf{RPE-CPM} \cite{he_generative_nodate}: An enhanced version of BART-B with \textit{reconstructed positional encoding (RPE) and constrained pointer mechanism (CPM)}. 
    
    \item \textbf{Span-Biaffine-ST} \cite{morio_end--end_2022}: Latest dependency parsing approach based upon Longformer \cite{Beltagy2020LongformerTL} along with \textit{Span-Biaffine} architecture for information sharing among sub-tasks. Since our approach is based upon a \textit{single-task setup (ST)}, we take the ST version of this work.
    
    \item \textbf{SP-AM} \cite{kawarada-etal-2024-argument}: Adapted TANL framework for \textit{Structured Prediction} for end-to-end AM based upon text-to-text generation.

    \item \textbf{DENIM} \cite{sun-etal-2024-discourse}: A discourse structure aware generative framework with multitask prompt-tuning.

\end{enumerate}

For the \texttt{AAE} benchmark, we evaluate the proposed method against all these baselines. For the \texttt{AAE-FG} dataset, we benchmark our results by running the most competitive open-source baseline \textbf{Span-Biaffine-ST}, using our implementation. For the \texttt{CDCP} benchmark, we use the following baselines: \textbf{BiPAM}, \textbf{BART-B}, \textbf{RPE-CPM}, \textbf{CPM-only} (without RPE)\cite{he_generative_nodate}, \textbf{SP-AM}, and \textbf{Span-Biaffine-ST}.


\subsection{Performance Metrics for Evaluation}
\label{sec:decode}
After generating the target ANL, we post-process the sequence by removing the symbol tokens to get a cleaned text. Finally, for a comprehensive evaluation, AC and AR tuples are created, including their types and corresponding boundaries. Following \cite{eger_neural_2017}, we evaluate the results with \textit{micro-F1} score for both ACE and ACRE tasks, where an exact match with the gold label is considered as a true label. 

\section{Results and Discussion}

In our discussion, we refer to the \textbf{\textit{arg}TANL} framework as $T5_{argTANL}$, \textbf{\textit{ME-arg}TANL} as $T5_{ME-argTANL}$, and the \textbf{\textit{Marker-Based Fine-Tuning}} as $T5_{MFS}$. Here, MFS indicates a specific Marker-Based Fine-Tuning Strategy: A-MKT, E-MKT, SM-MKT, or D-MKT.

Table \ref{baselines} compares the performance of the proposed frameworks with the baselines on the ACRE task in terms of micro F1-scores for component classification and relation classification. For the \texttt{AAE} dataset, $T5_{ME-argTANL}$ turns out to be the best among the proposed frameworks, surpassing \textit{arg}TANL and all the \textit{MFS-based} strategies, for component classification. While $T5_{ME-argTANL}$ outperforms several baselines for component and relation classifications, it gave a competitive performance to the current state-of-the-art models for component classification. We observe a similar performance on the \texttt{AAE-FG} dataset. For the \texttt{AAE-FG} dataset, $T5_{ME-argTANL}$ surpasses the current state-of-the-art result for component classification.

The likely reason behind the relatively poor performance of \textit{MFS-based} strategies is that models can detect markers within and outside the AC spans. However, in the second step of \textit{arg}TANL output generation, containing ACs and ARs, there is a high probability that some ACs might get split into smaller segments due to the presence of marker-like tokens within the AC spans. This erroneous splitting increases the number of potential permutations of ARs with these split ACs. Conversely, in the $T5_{ME-argTANL}$ model, both the markers and AC/AR labels are generated within the same target sequence. This allows the model to understand markers' relative positions and corresponding ACs. These markers act as contextual guides with the immediately preceding markers to automatically exclude markers inside the AC spans. This underscores the efficiency of the \textbf{\textit{ME-arg}TANL} by directly including argumentative markers within the target generation process, along with the augmented ACs and ARs, as opposed to employing any of \textit{MFS-based} technique.

D-MKT produces the best result for the \texttt{CDCP} dataset among the \textit{arg}TANL frameworks. Though it could outperform most of the baselines, it gave the second-best performance in terms of micro-F1 score for the component classification. It shows the strength of using the \textbf{\textit{MFS-based}} approach, where the model benefits from the contextual cues provided by DMs for identifying ACs. Across all the datasets, none of the proposed frameworks could beat the state-of-the-art result on relation classification within the joint formulation.

The above results indicate that the specific nature of argumentative markers, which are highly beneficial in \texttt{AAE} dataset, might not align as effectively with the argumentative structures present in the \texttt{CDCP} dataset. The A-MKT, SM-MKT, and E-MKT strategies are likely affected by \textit{catastrophic forgetting} \cite{Luo2023AnES}, a phenomenon where the model partially forgets previously learned knowledge during the target task fine-tuning. The poor performance of E-MKT across all datasets indicates that the knowledge of the relative position of markers is not beneficial. Similarly, the A-MKT strategy, which involves similar tasks set up in both steps of fine-tuning, proves to be sub-optimal. In contrast, the span-masking-based strategy SM-MKT demonstrates superior performance compared to both E-MKT and A-MKT.

\begin{table}
\caption{
Performance difference on the ACE task: Component-Only vs. End-to-End variants. Among all techniques, minimum differences between these two variants are marked in \textbf{bold}.
}
\centering
\renewcommand{\arraystretch}{1.1}
\scriptsize
\begin{tabular}{llccc}
\hline

\hline
\textbf{Corpus} & \textbf{Model} & \multicolumn{2}{c}{\textbf{Micro-F1 of ACE}} & \textbf{Diff. ($\pm$)} \\

 &  & \textbf{Comp-Only} & \textbf{End-to-End} &  \\
\hline
 & $T5_{argTANL}$ & 74.12 & 75.93 & -1.81\\
 & $T5_{ME-argTANL}$ & 74.48 & 77.14 & -2.66\\
 AAE & $T5_{E-MKT}$ & 71.64 & 73.06 & \textbf{-1.42}\\
 & $T5_{A-MKT}$ & 72.20 & 74.22 & -2.62\\
 & $T5_{SM-MKT}$ & 73.63 & 75.91 & -2.28\\
 & $T5_{D-MKT}$ & \textbf{74.87} & 76.45 & -1.58\\
\hline

 & $T5_{argTANL}$ & 59.20 & 60.41 & -1.21\\
 & $T5_{ME-argTANL}$ & 58.90 & 61.39 & -2.49\\
 AAE-FG & $T5_{E-MKT}$ & 55.15 & 55.24 & \textbf{-0.09}\\
 & $T5_{A-MKT}$ & 56.08 & 57.66 & -1.58\\
 & $T5_{SM-MKT}$ & 57.32 & 59.37 & -2.05\\
 & $T5_{D-MKT}$ & \textbf{59.29} & 59.71 & -0.42\\
\hline

& $T5_{argTANL}$ & 66.40 & 66.56 & -0.16\\
 & $T5_{E-MKT}$ & 59.05 & 57.31 & \textbf{+1.74}\\
 CDCP & $T5_{A-MKT}$ & 60.43 & 61.06 & -0.63\\
 & $T5_{SM-MKT}$ & 62.76 & 64.17 & -1.41\\
 & $T5_{D-MKT}$ & \textbf{66.66} & 67.49 & -0.83\\
\hline

\end{tabular}
\label{comp-only}
\end{table}

\begin{figure}
  \centering
  \begin{minipage}{0.24\textwidth}
    \centering
    \includegraphics[width=\textwidth]{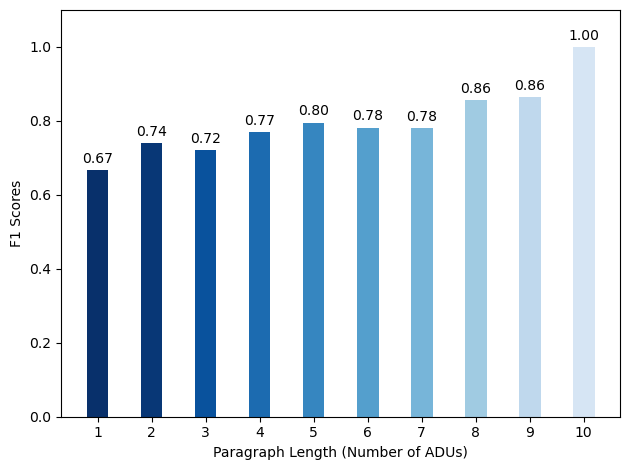}
  \end{minipage}
  \hfill
  \begin{minipage}{0.24\textwidth}
    \centering
    \includegraphics[width=\textwidth]{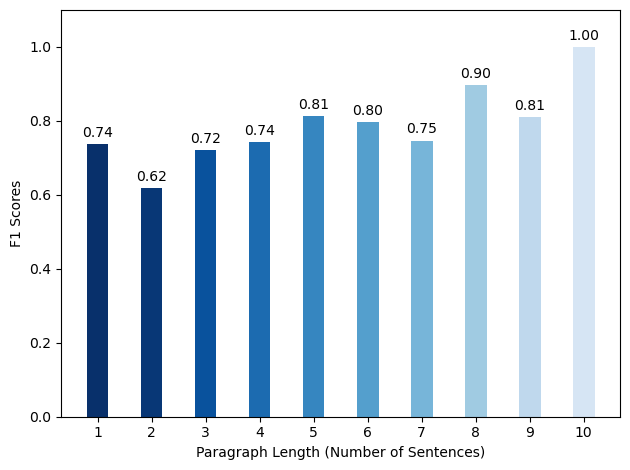}
  \end{minipage}
  \caption{Performance on the ACE task of the End-to-End variant with the varying number of ADUs (left) and of sentences (right) in a paragraph.}
  \label{fig:f1_ace}
\end{figure}

\subsection{Component Only vs End-to-End Variants}
Table \ref{comp-only} compares the performance of models on the ACE task in the two formulations. 
Analyzing the argumentative structure of \texttt{AAE} and \texttt{AAE-FG} datasets reveals that argument components always form a tree, indicating a single parent for each child. In this well-structured dataset, the performance of the models drops in the \textit{Component-only} variant compared to the \textit{end-to-end} variant. The \texttt{CDCP} dataset contains many isolated components. It does not necessarily form a tree but can form a graph (e.g., $t_1^c$ → $t_2^c$, $t_2^c$ → $t_3^c$, and a transitive relation $t_1^c$ → $t_3^c$). We observe the similar performance of models, except E-MKT, in the two variants for \texttt{CDCP} corpus. This result indicates that ACs and ARs benefit from mutual feature information, suggesting synergy in an end-to-end setup.
For \textit{Component-only} variant, the D-MKT strategy proves to be beneficial for all the datasets. When comparing the compatibility of the E-MKT with other \textit{MFS} strategies, it appears more suitable for the Component-only variant rather than the end-to-end for the ACE task. It exhibits the minimum performance difference in the Component-only setup for both versions of the \texttt{AAE} corpus and even shows performance improvements for the \texttt{CDCP} corpus.

\subsection{Performance Analysis Based on the Length of the Input}
We assess the performance of \textit{arg}TANL on the ACRE task for extracting ACs for the \texttt{AAE} dataset. Our analysis considers both the number of input text sentences and the number of ADUs. As illustrated in Figure \ref{fig:f1_ace}, the performance monotonically increases with increasing input length. This indicates that the performance of \textit{arg}TANL improves as the length of the paragraphs increases. 

\subsection{Ablation with the \textbf{Abbreviated \textit{arg}TANL} framework}

\begin{figure}
  \centering
  \includegraphics[width=0.45\textwidth]{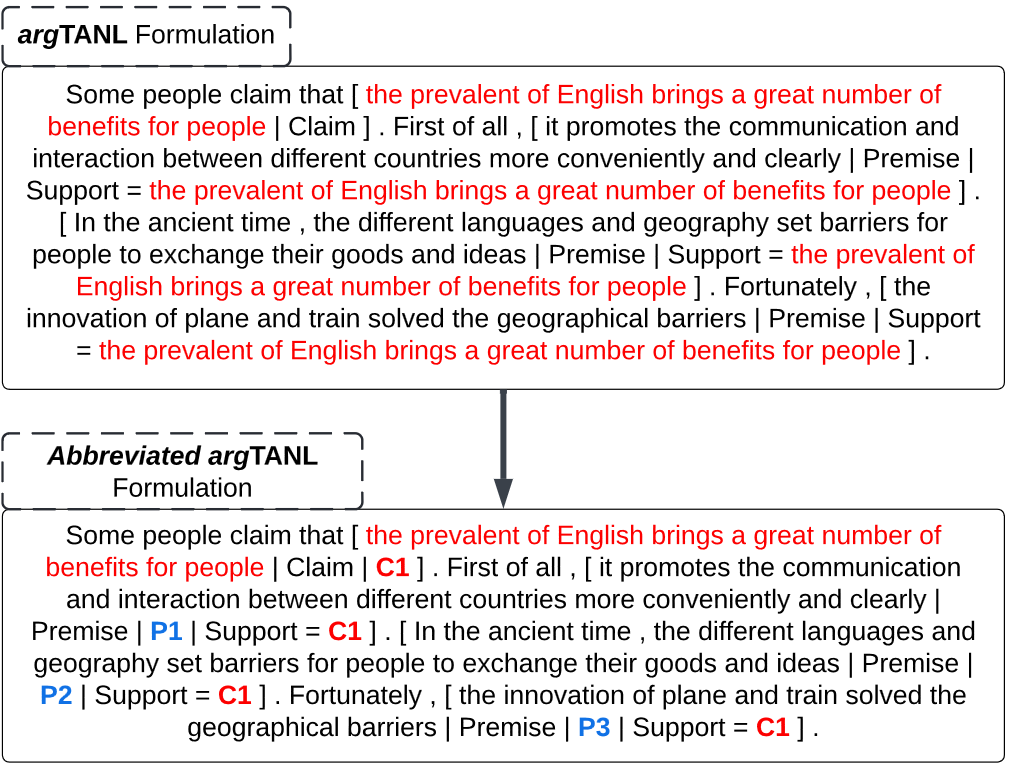}
  \caption{Representation of \textbf{Abbreviated \textit{arg}TANL} with \textit{``ID"} tokens replacing repeating spans for a shorter ANL output format.}
  \label{fig:abbreviated_formulation}
\end{figure}

This representation (See Figure \ref{fig:abbreviated_formulation}) is designed to minimize redundant and repeated AC spans connected to the right side of the \textit{Relation-type}. Here, we add an extra \textit{``ID"} token to every unique AC present in a paragraph in an ascending manner. For example, if $n$ number of ACs of type $Premise$ and $m$ number of $Claims$ are present in a paragraph, then, we will create abbreviated labels in the form of \textit{``ID"} tokens as \textit{P1, P2, P3...} for all $n$ number of \textit{Premises} and \textit{C1, C2, C3...} for all $m$ number of \textit{Claims}. These tokens will then be appended just after the original AC label as a \textit{``unique ID"} of the original label. Later, whenever the same AC is repeated during the relation assignment with \textit{Relation-type}, only the abbreviated tokens will be used for the same. This way we are keeping the related AC connections intact while reducing the size of the output sequence, allowing longer paragraphs to fit within the permissible output length.

\begin{table}
\renewcommand{\arraystretch}{1.1}
\centering
\caption{Performance comparison between \textit{\textit{arg}TANL} and \textit{Abbreviated argTANL} formulations.}
\scriptsize
\begin{tabular}{c|ccc|ccc}
  \hline
   & \multicolumn{3}{c|}{\textbf{Micro-F1 of ACE}} & \multicolumn{3}{c}{\textbf{Micro-F1 of ARC}}\\
  \hline
   \textbf{Dataset}  & \textbf{\textit{arg}TANL} & \textbf{Abbr.} & \textbf{Diff.} & \textbf{\textit{arg}TANL} & \textbf{Abbr.} & \textbf{Diff.} \\
  \hline
  AAE        & 75.93      & 75.48      & -0.45       & 50.56      & 47.32      & -3.24       \\
  AAE-FG     & 60.41      & 59.50      & -0.91       & 32.14      & 28.44      & -3.70       \\
  CDCP       & 64.78      & 65.16      & 0.38        & 20.65      & 18.20      & -2.45       \\
  \hline
\end{tabular}
\label{abbreviated_vs_vanilla}
\end{table}

Experiment results in Table \ref{abbreviated_vs_vanilla} suggest that, for the \texttt{AAE}, \texttt{AAE-FG}, and \texttt{CDCP} datasets, the \textit{Abbreviated} formulation achieved impressive average reductions in output sequence length of 23.56\%, 23.01\%, and 16.45\%, respectively. Also, this formulation resulted in competitive performance for the ACE task, showcasing its efficiency. While the ARC task experienced a noticeable decline in performance across all datasets with this formulation, the substantial reduction in output length represents a compelling trade-off. This suggests that this formulation effectively balances performance and efficiency, capturing essential information with shorter spans in the form of \textit{``IDs"}. However, for tasks that demand a deeper understanding of relational context, the \textit{arg}TANL framework, with its longer spans, remains superior. Overall, the \textit{Abbreviated} formulation presents pragmatic and resourceful insights, offering significant benefits in terms of output length reduction while maintaining competitive performance for certain tasks.

\subsection{Error Analysis}
\begin{table}
\caption{Error analysis for the ACRE task. \textbf{IT, IC,} and \textbf{IF} refer to \textit{Invalid Token, Invalid Component,} and \textit{Invalid Format} respectively.}
\centering
\renewcommand{\arraystretch}{1.1}
\scriptsize
\begin{tabular}{lllll}
\hline
\textbf{Corpus} & \textbf{Model} & \textbf{IT} & \textbf{IC} & \textbf{IF}\\
\hline
 & $T5_{argTANL}$ & 2.95 & 4.51 & 1.11\\
 & $T5_{E-MKT}$ & 5.45 & 6.62 & 3.39 \\
 AAE & $T5_{A-MKT}$ & 2.61 & 5.4 & 1.05 \\
 & $T5_{SM-MKT}$ & 3.06 & 5.23 & 1.39 \\
 & $T5_{D-MKT}$ & \textbf{2.39} & \textbf{4.9} & \textbf{0.8} \\
 & $T5_{ME-argTANL}$ & 4.95 & 5.57 & 1.00\\
\hline

 & $T5_{argTANL}$ & 3.23 & 6.07 & 1.05\\
 & $T5_{E-MKT}$ & 10.47 & 10.41 & 7.01 \\
 AAE-FG & $T5_{A-MKT}$ & 3.39 & 8.85 & \textbf{1.00} \\
 & $T5_{SM-MKT}$ & 3.45 & \textbf{5.29}& 1.33 \\
 & $T5_{D-MKT}$ & \textbf{2.95} & 5.73 & 1.22 \\
 & $T5_{ME-argTANL}$ & 5.45 & 7.29 & 1.05\\
\hline

 & $T5_{argTANL}$ & 11.33 & \textbf{4.93} & 7.6\\
 & $T5_{E-MKT}$ & 27.33 & 15.2 & \textbf{5.73} \\
 CDCP & $T5_{A-MKT}$ & 13.6 & 8.93 & 7.33 \\
 & $T5_{SM-MKT}$ & 12.53 & 6.4 & 7.2 \\
 & $T5_{D-MKT}$ & \textbf{9.6} & \textbf{4.93} & 7.06 \\
\hline


\end{tabular}
\label{errors}
\end{table}

The proposed method sometimes produces invalid ANLs as the generation is not fully controllable. We identified the following three major types of ANL-related erroneous generation (see Table \ref{errors}): \textit{(i) Invalid Token:} The generated ANL consists of some out-of-vocabulary tokens or out-of-context text spans (\textit{Hallucinations}). \textit{(ii) Invalid Format:} The invalid ANL format includes mismatched brackets, symbols, or corrupted text. \textit{(iii) Invalid Component:} The tail component connected with the relation in ANL is invalid if it is a span of text from the non-component regions. Importantly, these errors are not very frequent, and erroneous generations are discarded as negative results without undergoing any additional post-processing.
Among all the datasets, D-MKT is the most effective in generating error-free ANL, whereas E-MKT produces the highest number of errors for the ACRE task.

\section{Conclusion}
This work introduces proposes \textit{arg}TANL framework, a generative approach to AM. We demonstrate a text-to-text generation model could effectively handle the complexities of \textit{End-to-End} AM tasks. We further enhance the \textit{arg}TANL framework through \textit{Marker-Based Fine-Tuning}, which utilizes argumentative markers and DMs to improve the performance. This fine-tuning step enables the model to better recognize and extract ACs by learning the marker information. Additionally, we propose \textit{ME-arg}TANL framework, which directly integrates marker knowledge into the output sequence. \textit{ME-arg}TANL framework simultaneously identifies markers and argumentative structures. This dual-task approach yields superior performance compared to traditional methods. Through extensive experimentation on multiple AM benchmarks, we validate the proposed approaches \textit{arg}TANL, Marker-Based Fine-Tuned \textit{arg}TANL, and \textit{ME-arg}TANL by showcasing strong performance.
The ablation study on Abbreviated \textit{arg}TANL provides additional evidence of the efficiency and effectiveness of our framework, especially in managing longer sequences. Thus, our research highlights the potential of generative models in the AM domain. This also paves the way for future research in integrating domain-specific knowledge into generative models for complex NLP tasks.

\bibliography{main}
\bibliographystyle{IEEEtran}


 





\end{document}